# Structural features of the fly olfactory circuit mitigate the stability-plasticity dilemma in continual learning

Heming Zou[1†], Yunliang Zang[2†], Xiangyang Ji[1*]

[1]Department of Automation, Tsinghua University, Beijing, China
[2]Academy of Medical Engineering and Translational Medicine, Tianjin University, Tianjin, China
[†]These authors contribute equally to this work.
*Correspondence: xyji@tsinghua.edu.cn

**Abstract**

Artificial neural networks face the stability-plasticity dilemma in continual learning, while the brain can maintain memories and remain adaptable. However, the biological strategies for continual learning and their potential to inspire learning algorithms in neural networks are poorly understood. This study presents a minimal model of the fly olfactory circuit to investigate the biological strategies that support continual odor learning. We introduce the fly olfactory circuit as a plug-and-play component, termed the Fly Model, which can integrate with modern machine learning methods to address this dilemma. Our findings demonstrate that the Fly Model enhances both memory stability and learning plasticity, overcoming the limitations of current continual learning strategies. We validated its effectiveness across various challenging continual learning scenarios using commonly used datasets. The fly olfactory system serves as an elegant biological circuit for lifelong learning, offering a module that enhances continual learning with minimal additional computational cost for machine learning.

Despite significant progress across many fields, artificial neural networks (ANNs) still suffer from the stability-plasticity dilemma (Fig. 1a). When learning new tasks and updating parameters, these models inevitably overwrite previously learned patterns, resulting in "catastrophic forgetting" [1-3]. This critical flaw has become the Achilles' heel of neural network models, preventing them from realizing their full potential. Conversely, especially under long non-stationary data streams, the parameters of network models may become less effective at updating, resulting in a gradual decline in their ability to adapt to new information. This issue, known as plasticity loss, has garnered increasing attention in recent years [4-5].

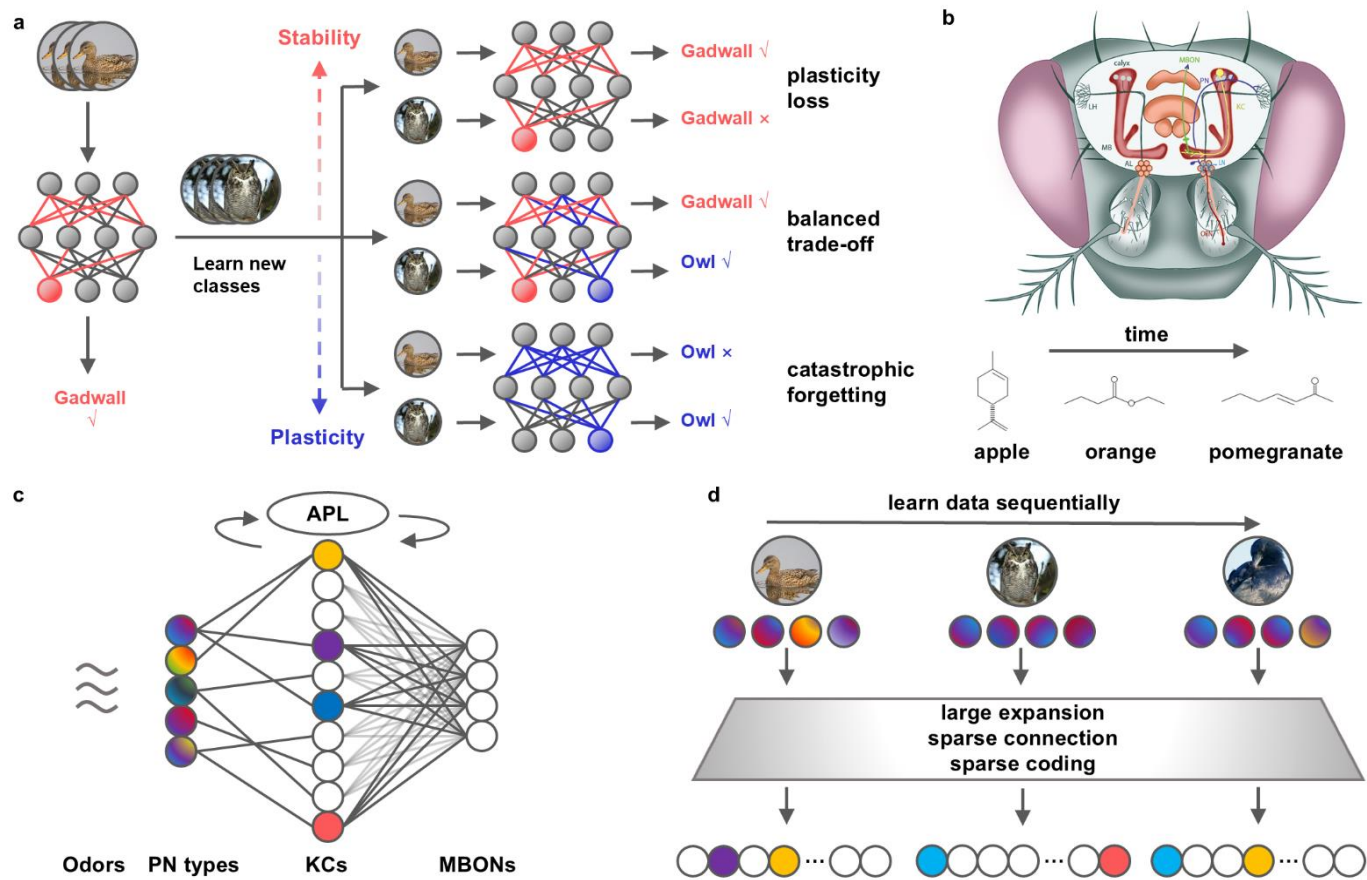

**Fig. 1 Schematic of how the fly olfactory circuit features can alleviate the stability-plasticity dilemma. a** ANNs often encounter issues of catastrophic forgetting and plasticity loss during continual learning. **b** Sequential learning of distinct odors in the fruit fly olfactory circuit. **c** Schematic of the three-layer fly olfactory circuit model, including PNs, KCs, and MBONs. There are more KCs than PNs, and the synaptic connections between PNs and KCs are random and sparse. APL neurons exert strong inhibition, implementing a winner-take-all strategy. **d** The fly olfactory circuit may enhance continual learning by its critical features.

Many strategies have been developed to facilitate continual learning (CL) by balancing the stability and plasticity of network models. Existing approaches to alleviate catastrophic forgetting include regularization-based [6-10], ensemble-based [11-17], and replay-based methods [18-23]. To mitigate plasticity loss, researchers have proposed resetting-based [5, 24-27], regularization-based [28-31], architecture-based [4, 33], and optimizer-based strategies [4]. However, these methods primarily focus on enhancing one aspect—either stability or plasticity. Few studies attempted to address both issues simultaneously by balancing the trade-off between the two factors [34-37]. In addition to common issues such as excessive computational costs and lack of biological plausibility, each method has its own challenges. For instance, they may require access to previous samples [34], introduce auxiliary networks [35], or exhibit a significant performance gap compared to the latest methods focused on forgetting or plasticity [36-37] (see also the discussion in the Supplementary material).

In contrast to ANNs, the brain maintains a good balance between plasticity and stability during CL, without exhibiting severe catastrophic forgetting or plasticity loss (Fig. 1b). While it is widely accepted that neural connection patterns enable neurons to form specialized circuits for specific functions, thereby enhancing an organism's learning capabilities [38], it remains unclear how a specific circuit mitigates or overcomes the stability-plasticity dilemma by leveraging various biological strategies in CL. Additionally, studying how biological organisms address this dilemma can inspire the design of more efficient neural networks and learning algorithms to facilitate CL.

Due to its anatomical and computational simplicity (Fig. 1b, c), the fly olfactory circuit has been extensively used to explore the neural principles underlying information processing [39-42]. It is not surprising that flies learn different odors sequentially rather than being exposed to all odors simultaneously. In this circuit, odor-evoked responses from olfactory receptor neurons first propagate through approximately 50 glomeruli, then to projection neurons (PNs), and finally to around 2000 Kenyon cells (KCs) before reaching the mushroom body output neurons (MBONs) to select an action. Three main features of this circuit critical for information representation and pattern classification include (Fig. 1d): (1) an expansion ratio of approximately 40 from PNs to KCs (Feature 1—large expansion); (2) each KC forms synapses with approximately 6 PNs [42] (termed synaptic degree, Feature 2—sparse connections); and (3) a feedback inhibition loop formed by KCs → Anterior paired lateral (APL) neurons → KCs, allowing only the KCs with the strongest excitation to fire (Feature 3—sparse coding).

In this work, we hypothesize that the simple yet elegant features of the fly olfactory circuit—large expansion, sparse connection, and sparse coding—play critical roles in CL in biological models (Fig. 1d) and can help address the stability-plasticity dilemma in machine learning (ML) models with low additional computational costs. To test this hypothesis, we first built a minimal model that captures the key features of the fly olfactory circuit to evaluate its ability to continuously learn different olfactory stimuli and analyze the importance of each feature. Subsequently, we propose the fly olfactory circuit as a plug-and-play module to integrate with modern ML methods to improve both stability and plasticity under the widely used Class Incremental Learning (CIL) setup. Various datasets, including CIFAR-100 [43], CUB-200-2011 [44], and VTAB [45], were used in our simulations. The results demonstrate that the module based on the fly olfactory circuit efficiently improves CL performance, enhancing existing strategies for mitigating forgetting and plasticity loss with minimal computational overhead. We also examined the module's effectiveness and robustness in the context of streaming learning and class-imbalanced scenarios.

## Results

**CL in the fly olfactory circuit.** Our first goal was to build a minimal model of the fly olfactory circuit to simulate its ability to learn distinct odors sequentially (Fig. 1c). The neuronal connections between the PN layer and the KC layer are determined by a random sparse expansion projection matrix [42]. Similar to previous work, each neuron in the KC layer is connected to $r$ PNs (where $r$ is referred to as synaptic degree [46]), with the synaptic weight set to 1, while the other weights in the connection matrix are set to 0, indicating no connection. For KCs, we apply the top-$k$ operation to simulate the strong inhibition from APL neurons, activating only the top $k$ percent of neurons with the strongest odor stimulation (winner-take-all). We denote $k$ as the coding level in neuroscience [47-48]. Subsequently, KCs are fully connected to MBONs that represent different odor categories. Consistent with previous biological models [42], only the synaptic weights between the KC and MBON layers are updatable, while the connections between the PN and KC layers remain frozen during learning. The parameter update mechanism in the fly olfactory circuit is still an open question in neuroscience. Here, we adopted the efficient Synaptic Intelligence (SI) algorithm to update synaptic weights, an effective CL method designed to maintain memories for older tasks [9]. For comparison, we also utilized the widely used Stochastic Gradient Descent (SGD) method, which aims to minimize the loss function by updating synaptic weights without specifically considering the stability or plasticity of the network. Similar to previous work [46, 49], we simulated odors of corresponding categories by adding Gaussian-distributed noise to randomly selected class prototypes (class centroids). The dimensionality of the odor features is 50, the same as the number of glomeruli, through which signals are fed to PNs (see Methods for details). Following the setup of the CIL paradigm, we simulated 10 odor categories divided into 5 tasks for CL, with two new odors learned in each task (Fig. 2a).

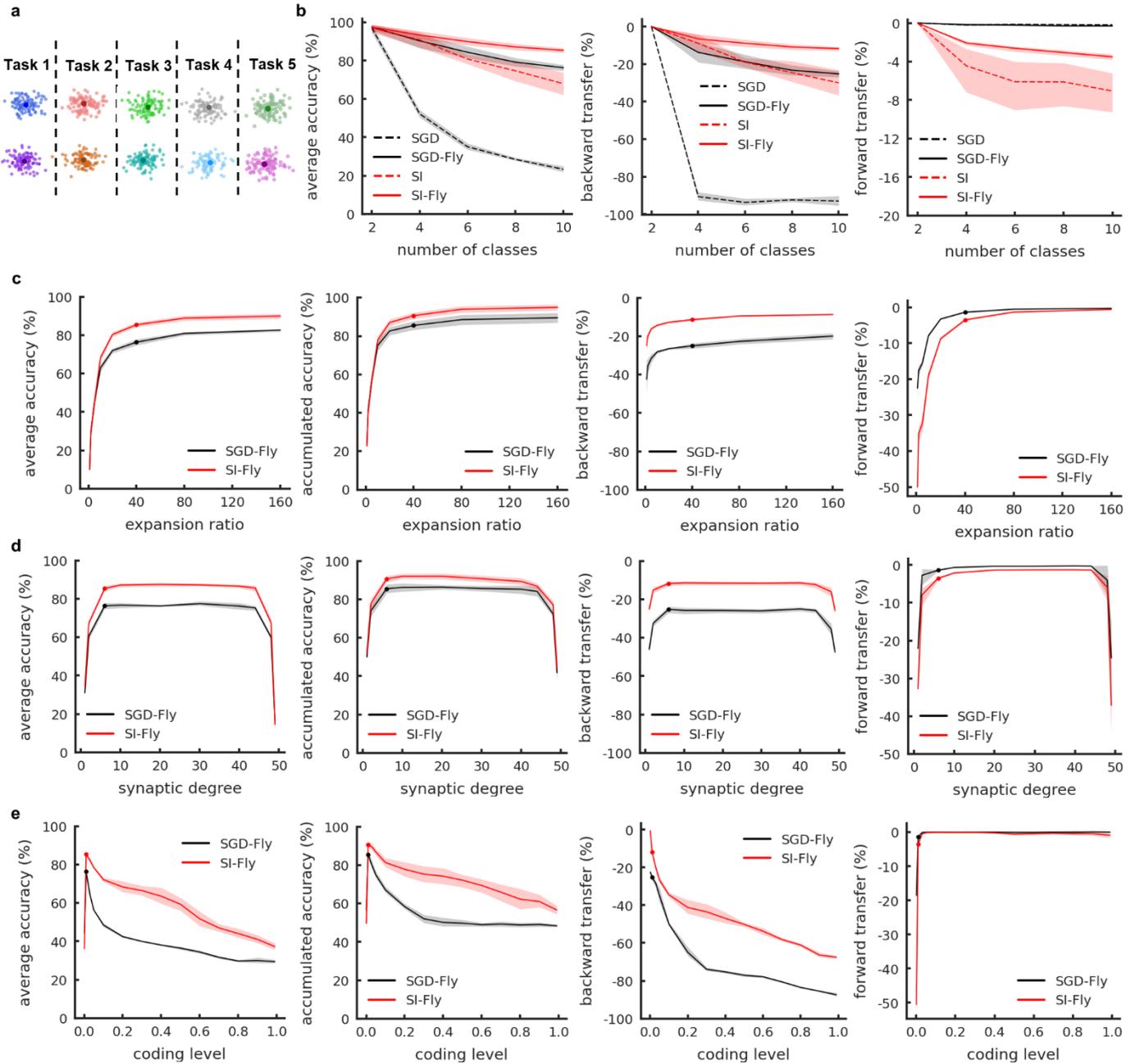

**Fig. 2 Sequential learning of odors in fly olfactory circuit models. a** Schematic of the sequential learning task. **b** Average accuracy, backward transfer, and forward transfer against the number of odor classes during CL. **c-e** Sensitivity of performance (average accuracy, accumulated accuracy, backward transfer, and forward transfer, from left to right) to expansion ratio, synaptic degree, and coding level, respectively. The dots denote the parameters used in Fig. 2b. In all plots, SGD (SI) refers to a two-layer network without KCs, with synaptic weights updated using the SGD (SI) method. SGD (SI)-fly refers to the baseline olfactory circuit with synaptic weights updated using the SGD (SI) method.

In Fig. 2b, we defined a baseline fly olfactory circuit model, with expansion ratio of being 40; the synaptic degree $r$, set to 6; and the coding level $k$, set to 1%. under this expansion ratio, we selected $r$ and $k$ as the hyperparameters that yield the best accuracy in this setting. Note that except the slightly lower $k$ value (experimentally 5%), all other parameters are identical to experimentally reported values [39-41, 50-51]. We used average accuracy to evaluate the performance after learning each task, backward transfer to measure forgetting, and forward transfer to measure plasticity loss (see Methods for details).

The simulation results indicate that the baseline fly olfactory circuit model maintains high accuracy even after learning 10 classes of odors sequentially, regardless of the synaptic update methods (SGD-fly and SI-fly in Fig. 2b left). Nonetheless, once ablating the fly olfactory circuit (SGD and SI in Fig. 2b by removing the KCs), the test accuracy significantly drops. By comparison, CL achieves a higher performance if SI is adopted as the synaptic updating method compared to SGD. From a fine-grained perspective, backward transfer and forward transfer can be considered as the components of accuracy decrease projected onto the dimensions of forgetting and plasticity loss, within the range of (-100%, 0). SI efficiently mitigates forgetting by using regularization to consolidate synaptic weights vital to previous tasks, which unavoidably leads to loss of plasticity as a side effect (Fig. 2b middle & right). In contrast, SGD exhibits almost no plasticity loss, but significant forgetting occurs during the process of CL. The fly olfactory model efficiently mitigates both forgetting and loss of plasticity when they exist. All these results support our hypothesis that the fly olfactory circuit supports CL of distinct odors.

In the next step, we analyzed the sensitivity of learning performance to the parameters that determine the three circuit features (Fig. 2c-e). We reported the average accuracy, accumulated accuracy (mean value of average accuracy before the current stage), as well as the last stage backward and forward transfer. The effects of the circuit parameters on learning performance are independent of the synaptic update rules. The expansion ratio facilitates CL by reducing both stability (backward transfer) and plasticity (forward transfer) loss, with the facilitation effect gradually saturating when the expansion ratio exceeds 40 (Fig. 2c). Both stability and plasticity are largely independent of the synaptic degree in the middle range but drop significantly when the number approaches either maximum or minimum values, leading to a similar trend in CL performance (Fig. 2d). The coding level in KCs is critical for CL. Memory stability decreases significantly with higher coding levels, but plasticity increases steeply in the initial range and then saturates. The net effect of these two factors is that there exists an optimal coding range for CL performance (Fig. 2e).

**Biological factors breaking the stability-plasticity dilemma in the fly olfactory circuit.** In this section, we outlined how the fly olfactory circuit addresses the stability-plasticity dilemma during CL of distinct odors.

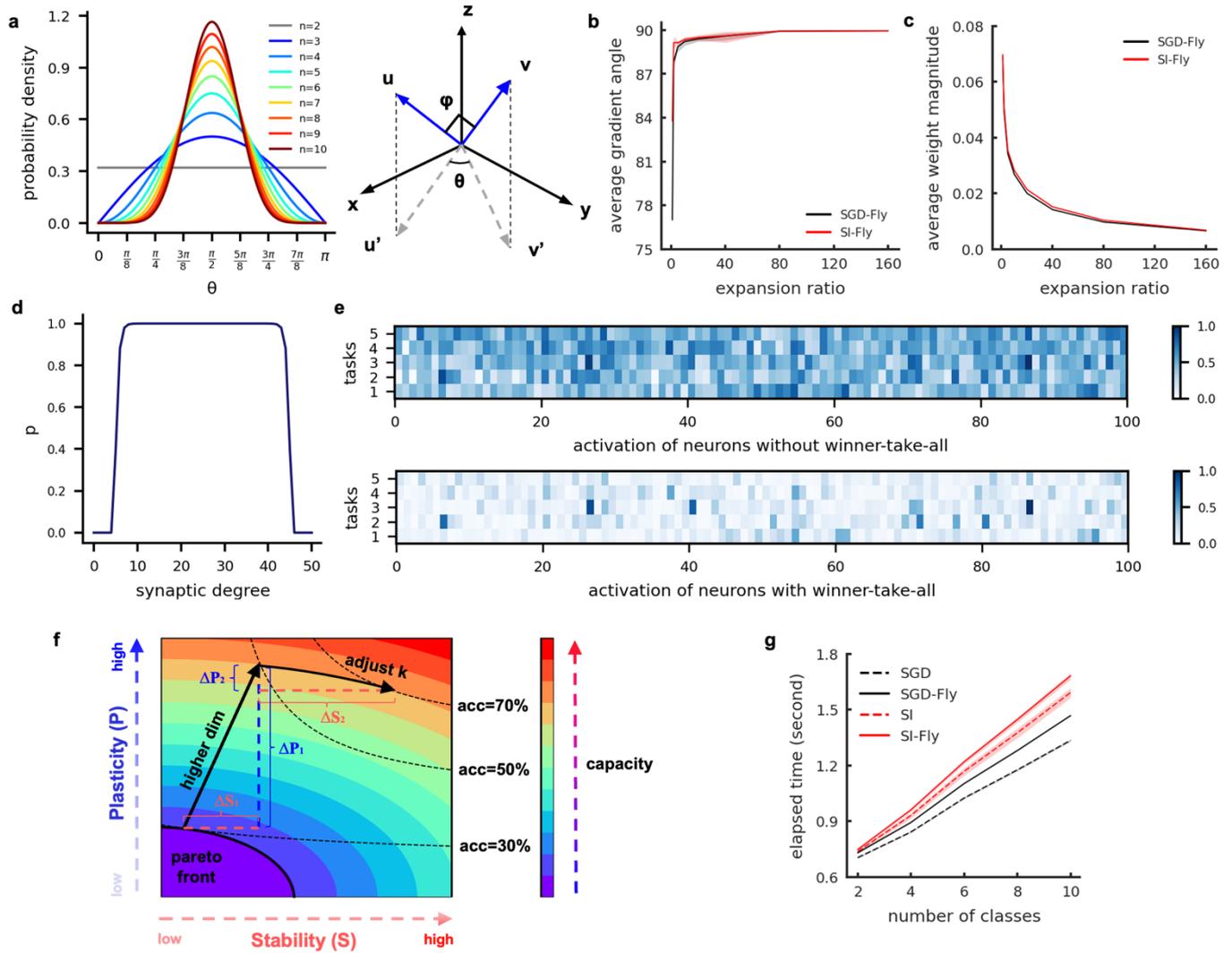

**Fig. 3 Biological factors facilitating CL in the fly olfactory circuit. a** Probability density of angle distributions between random vectors in different dimensions (left). Vectors are more likely to become orthogonal when projected to higher dimensions (right). **b** Average angle between gradients at the optima of tasks 1 ($\nabla L_1(w_1^*)$, after learning 2 classes) and 5 ($\nabla L_5(w_5^*)$, after learning 10 classes) with increasing expansion ratios. **c** Average magnitude of synaptic weights with increasing expansion ratios. In b-c, black and red traces indicate SGD and SI strategies, respectively. **d** Probability of each KC receiving distinct inputs against synaptic degree. **e** KC activation in different tasks (100 out of 2000 cells) without (top) and with (bottom) top-$k$ operation. **f** Overall effect of the fly olfactory model on CL. **g** Computational cost for SGD (SI) and SGD (SI)-Fly models.

Fig. 3a illustrates the probability density function of the angle distributions between two random vectors. When vectors are projected from a lower dimensional space to a higher dimensional space, they are more likely to be perpendicular, with reduced correlations. In the feedforward path, odor information is transformed by the fixed PN-to-KC projection matrix ($W_{P \to K}$) and the adaptable KC-to-MBON projection matrix ($W_{K \to M}$). The numerous KCs help make vectors representing different odors more orthogonal after passing through $W_{P \to K}$, reducing task interference and minimizing forgetting. For $W_{K \to M}$, we measured the angle between the gradients at the optima of task 1 ($\nabla L_1(w_1^*)$) and task 5 ($\nabla L_5(w_5^*)$) as shown in Fig. 3b. As the expansion ratio increases, this angle approaches 90 degrees, which helps prevent forgetting. When gradients for different tasks ($\nabla L_i$ and $\nabla L_j$) become more orthogonal, changes in weights for new tasks interfere less with the loss function of previous tasks [52] (see Theorem 1).

To our knowledge, there is no direct mathematical tool to prove that expansion enhances plasticity. We calculated the average magnitude of synaptic weights, a common metric for measuring plasticity (see Methods for details). Larger expansions result in decreased average magnitude of synaptic weights (Fig. 3c). Smaller network weights reduce the condition number of the Hessian matrix, accelerating the convergence speed of the SGD algorithm and aiding the network in finding a global maximum, thereby enhancing plasticity [53]. Another intuitive aspect is that more KCs increase the network's representational capacity, aiding adaptation to new tasks and consequently improving plasticity.

The synaptic degree is linked to the "effective dimension" of the KC layer. If two KCs are connected to identical PNs, their outputs are indistinguishable, leading to fewer effective KCs. Define *p* as the probability that each KC receives a distinct subset of inputs (Fig. 3d). The maximum probability *p* and the largest number of effective KCs occur when the synaptic degree is *n*/2 (where *n* is the number of glomeruli and PN types) [46]. Extremely small or large synaptic degrees reduce the number of effective KCs and dimensionality, causing declines in both stability and plasticity (shifting to the left in Figs. 3b-c).

Fig. 3e shows KC activation for different tasks. After the top-*k* operation (winner-take-all via strong APL inhibition), KC activity becomes sparse and well-separated, minimizing overlap. This creates isolated sub-networks for each task, enhancing stability. However, only a subset of neurons is active, leaving many dormant, which reduces network capacity and decreases plasticity. Consequently, backward transfer increases with sparser activity, while forward transfer decreases.

In Fig. 3f, the Pareto front illustrates the stability-plasticity dilemma, with each front representing the same representational capacity. The black dashed "iso-accuracy line" indicates a set of stability-plasticity combinations that achieve identical learning performance. Increasing the effective dimension (via larger expansions or synaptic connections) advances the Pareto front, enhancing both stability ($\Delta S_1$) and plasticity ($\Delta P_1$) simultaneously. Along this new boundary, an optimal coding level exists that balances stability ($\Delta S_2$) and plasticity ($\Delta P_2$), resulting in optimal learning performance. While increasing the expansion ratio and synaptic degree (optimal value ~25) consistently improves CL, these benefits come with higher metabolic and structural costs. Thus, the expansion ratio (~ 40) and synaptic degree (~ 6) in the fly olfactory circuit may represent a biological configuration that achieves effective performance with limited resources.

From a ML perspective, this lightweight circuit provides a computationally efficient framework. For a single sample, its binary and sparse structure reduces the Floating-Point Operations (FLOPs) from 100×2000 (with half additions and half multiplications) to 6×2000 (only additions) when mapping the 50-dimensional PNs to 2000-dimensional KCs, compared to a dense matrix. To update the synaptic weights between KCs and MBONs, only 6×20×10 FLOPs are needed (with a coding level of 0.01), which is about 1% of the FLOPs required to update all parameters. In practical tests (see Fig. 3g), the SGD (SI)-Fly incurs minimal additional time compared to SGD (SI).

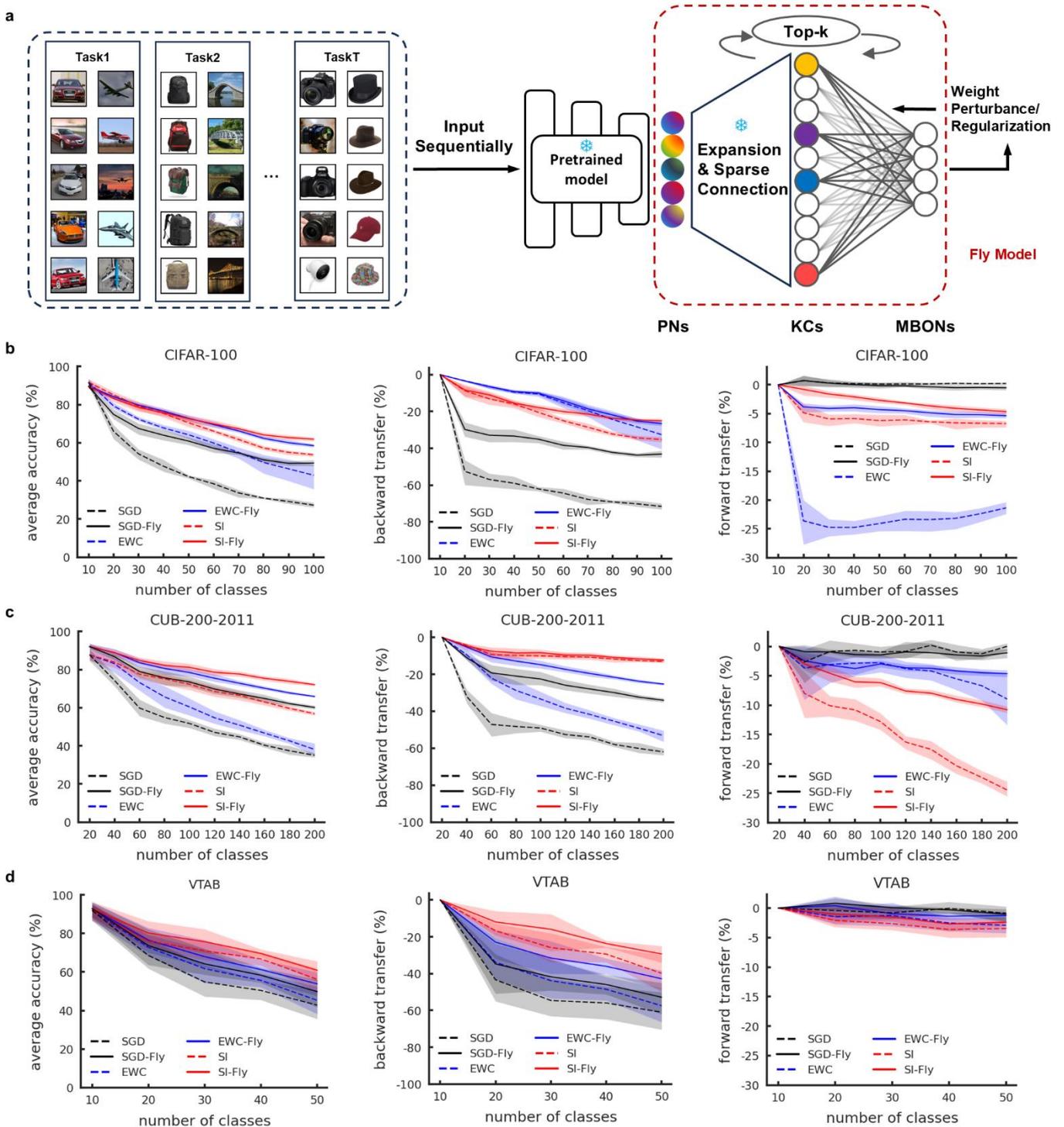

**Fig.4 The Fly Model improves stability and plasticity for stability-targeted methods. a** Model pipeline. **b** The average accuracy (left), backward transfer (middle), and forward transfer (right) against the number of classes learned using CIFAR-100. SGD-Fly, EWC-Fly, and SI-Fly refer to the backbone plus Fly Model using different synaptic updating methods (SGD, EWC, and SI). SGD, EWC, and SI correspond to the backbone plus the baseline Fly Model minus the KC layer with different synaptic update rules. **c, d** Same as **b**, but for CUB-200-2011 and VTAB, respectively.

**The Fly Model enhances stability and plasticity when combined with stability-targeted strategies in ML.** Based on the simulation results and analysis, we hypothesized that a model structure mimicking the fly olfactory circuit can facilitate CL in ML models. We proposed the fly olfactory circuit as a lightweight plug-and-play module, referred to as the Fly Model, and tested its effects on CL by integrating it with modern ML methods in the context of CIL.

In all subsequent ML tasks, CL performance was evaluated using CIFAR-100 (100 classes) [43], CUB-200-2011 (200 classes) [44], and VTAB (50 classes) [45], with 10, 20, and 10 classes per simulation task, respectively. Fig. 4a illustrates the pipeline of the first batch of simulations. We used the widely adopted pre-trained Vision Transformer [54] model as the backbone, extracting 768-dimensional image feature vectors to feed into the Fly Model. In the Fly Model, there are 768 PNs (matching the feature vector size), 30,000 KCs, and MBONs corresponding to the total class numbers: 100 for CIFAR-100, 200 for CUB-200-2011, and 50 for VTAB. The pre-trained model and PN→KC connections remain frozen during training, with only the connection weights between KCs and MBONs updated. We employed the commonly used SGD, elastic weight consolidation (EWC) [6], and SI [9] strategies to update the connection weights. EWC and SI are strategies aimed at enhancing stability.

The Fly Model's impact on CL is evident through consistent improvements in test accuracy across all datasets, regardless of the weight update methods used (Fig. 4b-d, left). The Fly Model increases test accuracy by 22% for CIFAR-100, 25% for CUB-200-2011, and 7% for VTAB when using SGD to update synaptic weights. The according accuracy improvements are 8%, 15%, and 5% with the SI method. Improvements fall within the middle range when using EWC. By analyzing test accuracy reduction into backward and forward transfer, we illustrated the Fly Model's effects on stability and plasticity changes during CL (Fig. 4b-d, middle & right). Across datasets, SGD causes minimal plasticity loss but experiences significant forgetting (stability loss) due to a lack of protection for previously learned synaptic weights. SI and EWC prevent forgetting by regularizing and consolidating important synaptic weights, but at the cost of reduced plasticity for new tasks. The Fly Model mitigates both stability and plasticity loss for stability-targeted SI and EWC. For SGD, the facilitation effect is mainly due to reduced overwriting, as SGD causes minimal plasticity loss.

These results indicate that neural networks often suffer more from forgetting than plasticity loss in CIL experiments, highlighting where the Fly Model contributes most in this context. However, plasticity loss can become more severe in longer non-stationary sequence learning (see Fig. 6).

**The Fly Model improves stability after combining with strategies targeted at enhancing plasticity.** In the previous section, the facilitation effects of the Fly Model on CL have been mainly tested on strategies that mitigate forgetting. There also exist strategies aimed at alleviating plasticity loss [28-29]. In Fig. 5, we tested the facilitation effects of the Fly Model by combining it with two recent representative methods, L2 Init [29] and Shrink and Perturb (S&P) [28]. These methods incorporate either a regularization term on the initial parameters or parameter perturbations to enhance plasticity. The commonly used SGD method is preserved for comparison.

For plasticity-targeted methods, we employed a grid search to identify hyperparameters that yield the best test accuracy. Across all datasets, the CL performance of L2 Init and S&P is indistinguishable from that of the SGD method. In the current settings, forgetting is the primary issue, while plasticity loss is less significant. The test accuracy, along with backward transfer, decreases when hyperparameters aimed at improving plasticity are applied (Supplementary Fig. 1), suggesting that plasticity-targeted methods tend to sacrifice stability for enhanced plasticity. Optimal performance is achieved when these hyperparameters are close to zero, effectively reverting to basic SGD. These results indicate that plasticity-targeted methods do not alleviate catastrophic forgetting. However, the Fly Model significantly improves the ability of these methods to mitigate forgetting without causing plasticity loss.

The effectiveness and robustness of the Fly model were also tested in multilayer neural networks (where the recent continual backpropagation methods can be applied [5]) and class-imbalanced scenarios (Supplementary Figs. 2-3). While we could not exhaust all synaptic update strategies and datasets, the simulation results support our hypothesis that the fly olfactory model can, in principle, integrate with other ML strategies to enhance CL performance.

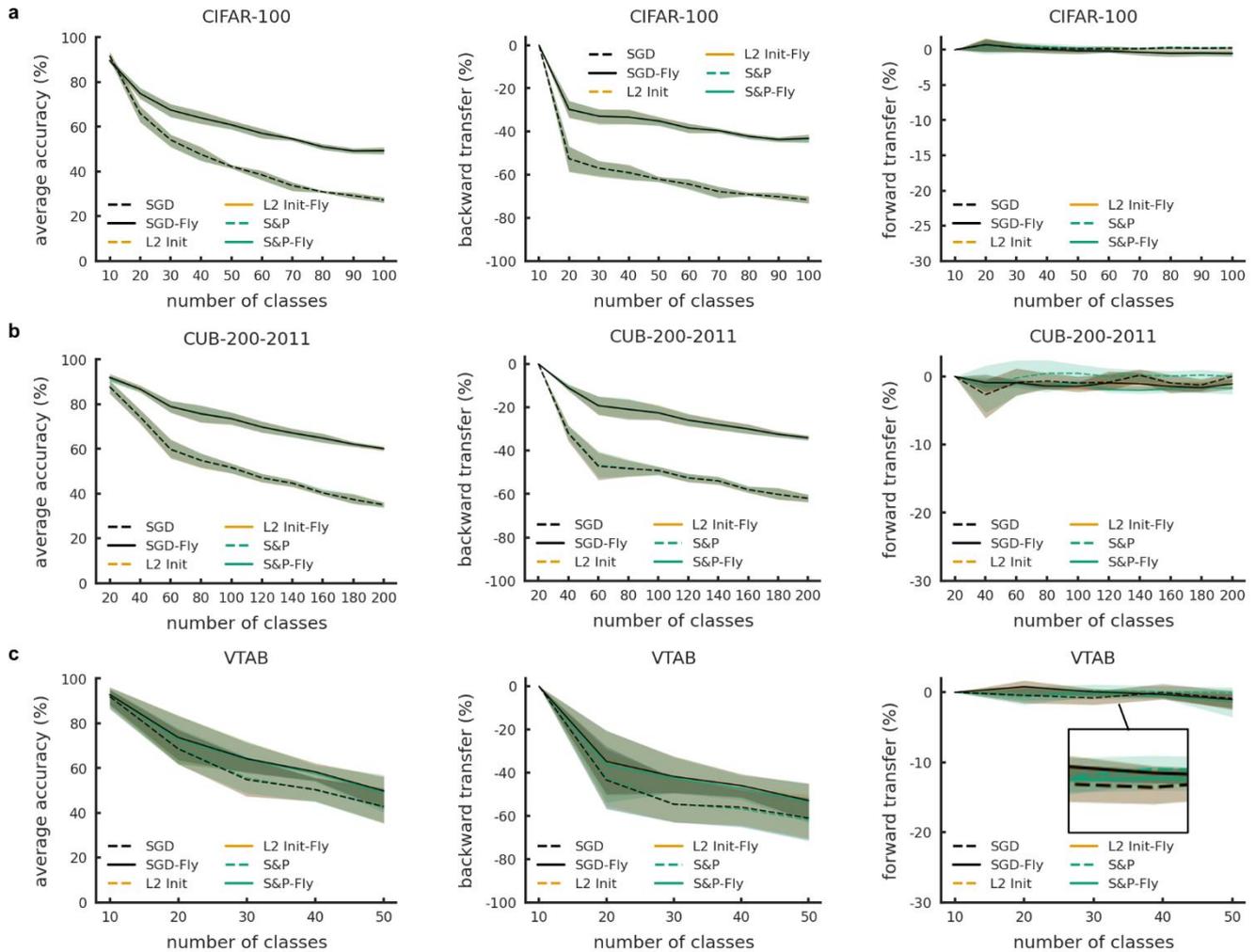

**Fig.5 The Fly Model improves stability for plasticity-targeted methods. a** The average accuracy (left), backward transfer (middle), and forward transfer (right) against the number of classes learned using CIFAR-100. SGD-Fly, L2 Init-Fly, and S&P-Fly refer to the backbone plus the Fly Model using different plasticity-targeted methods (SGD, L2 Init, and S&P). SGD, L2 Init, and S&P correspond to the backbone plus the baseline Fly Model minus the KC layer with different plasticity-targeted methods. **b, c** Same as **a**, but on CUB-200-2011 and VTAB, respectively.

**The Fly Model mitigates plasticity loss in long-sequence, non-stationary data streams.** In our CIL simulations, plasticity loss is less significant than stability loss, which limits the Fly Model's effectiveness in addressing plasticity loss. Therefore, we simulated streaming learning, where new information arrives in non-stationary online data streams without clearly defined task boundaries, leading to severe plasticity loss in neural networks.

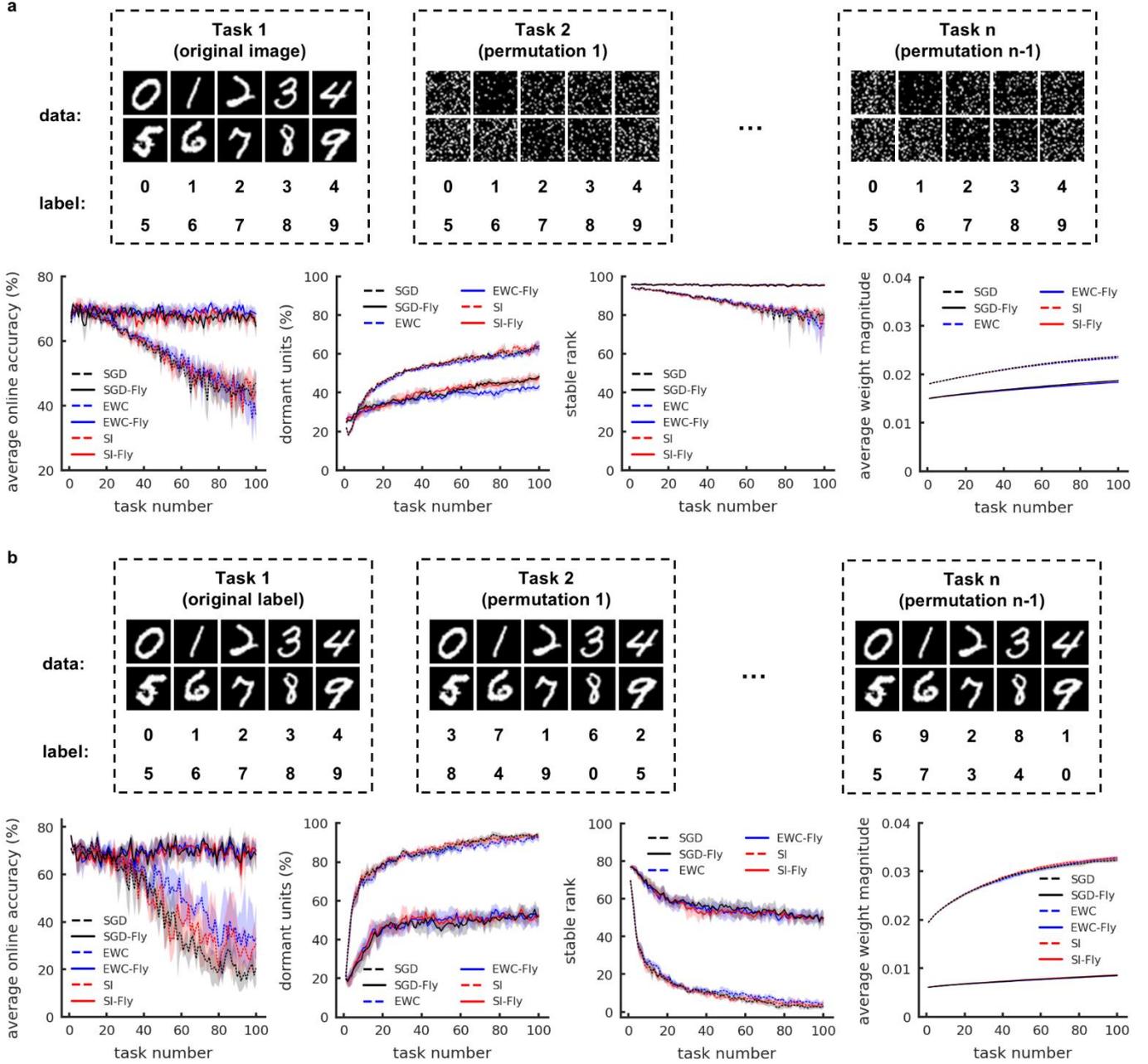

**Fig. 6 The Fly Model enhances stability-targeted methods in streaming learning. a** Streaming learning using Input Permuted MNIST. Metrics include average online accuracy, dormant units, stable rank, and average weight magnitude. The effects of the Fly Model are tested against synaptic update strategies of SGD, EWC, and SI. **b** Same as **a**, but using Label Permuted MNIST (Layout inspired by [1]).

Following previous studies [29, 36], we adopted two typical datasets: Input Permuted MNIST [55] and Label Permuted MNIST [29]. We generated 10,000 samples per task, with a total of 100 tasks in sequence. Each sample was fed into the network once, in an online learning manner. Throughout the 100 tasks, the network was trained to classify 10 classes. Unlike CIL, the number of classes does not increase over time. For Input Permuted MNIST, the input distribution shifts during CL, with the pixels of the same images randomly permuted in different orders for new tasks (Fig. 6a). For Label Permuted MNIST, concepts shift, where the same inputs are assigned randomly permuted labels for different tasks (Fig. 6b).

To evaluate the learning process from various perspectives, we employed four key metrics. Average online accuracy, a variant of average accuracy, measures plasticity. Dormant units indicate the ratio of neurons with nearly zero activation values; more dormant units suggest fewer active units available for adapting to new information. Stable rank, akin to effective rank in matrix theory, also measures plasticity, with a higher stable rank indicating greater plasticity [56]. Lastly, average weight magnitude is considered; according to ML theory, networks with lower weight magnitudes tend to exhibit higher plasticity [53] (see Methods for details).

Plasticity-targeted methods have already shown exceptional performance in this scenario, resulting in minimal plasticity loss. Therefore, we examined the Fly Model's impact on reducing plasticity loss in stability-targeted methods, using basic SGD for comparison. Simulation results reveal that EWC and SI perform similarly to basic SGD, experiencing significant plasticity loss. However, incorporating the Fly Model effectively mitigates this loss. Metrics such as average online accuracy (increased), dormant units (reduced), stable rank (increased), and average weight magnitude (reduced) all demonstrate consistent trends.

## Discussion

In recent years, brain-inspired computation has shown great promise in addressing AI challenges. Many studies draw inspiration from different brain circuits, integrating features like synaptic mechanisms, neuronal properties, and coding strategies to pursue human-like intelligence. However, not all features are optimal for computational and learning performance. Each feature may represent a trade-off tailored to specific functions, considering factors like metabolic costs, resilience, and learning speed. The fly olfactory circuit exemplifies this idea.

In CL, increasing the number of KCs and forming more synaptic connections with glomeruli can improve performance, but the marginal gains are offset by higher metabolic and structural costs. Maintaining memory stability requires avoiding the overwriting of learned synaptic weights, leading to the stability-plasticity dilemma. Previous studies have tried to balance these factors by efficiently searching for unused synaptic weights for new tasks while preserving important ones. For instance, smaller synaptic weights may help find solutions for new tasks by accelerating convergence [53]. However, they can't resolve the dilemma if the capacity limit is reached. Similar to neurogenesis in biological circuits [57], one simple solution in ML is to increase network capacity by adding neurons, which unavoidably increases computational costs.

The Fly Model elegantly overcomes this dilemma with three key features: large expansion, sparse connections, and sparse coding. It enhances CL in two ways. First, sparse coding in KCs balances stability and plasticity. Second, these features elevate the network's Pareto front for stability-plasticity with minimal additional computational costs (Fig. 3). Biologically, similar structures exist in the cerebellum, hippocampus, and electric fish, related to cerebellum-like circuits [39-40, 58-59].

In this work, we demonstrated that the three critical features of cerebellum-like circuits—large expansion, sparse connections, and sparse coding—can function as a plug-and-play component that integrates seamlessly with modern ML methods to reliably facilitate CL. Furthermore, the proposed Fly Model, a specific instance of cerebellum-like modules, may have broad applications in other ML tasks such as spatial navigation, anomaly detection, and language processing, owing to its strengths in feature extraction [59-60].

To make the Fly Model more practical for real-world applications, there is still room for improvement. The fly olfactory circuit model is minimal. More biologically efficient factors, such as branch-dependent computing [61], state-dependent learning [62-63], and multisite plasticity [64], have not been included in our model framework due to limited knowledge. These factors may be the missing pieces for CL. We have shown that there is an optimal coding level for any fixed scenario of CL, but this level can vary under different setups and network architectures. Developing an automatic parameter selection algorithm would greatly enhance real-world deployment. Furthermore, the sparsity property of the Fly Model has the potential to save significant resources, which is crucial for edge devices with limited computational capacities. If a specific operator could be implemented in hardware, it would further reduce computational consumption. Finally, while we tested the Fly Model in the context of image classification tasks, more general CL paradigms involving diverse cognitive tasks need to be explored. These represent promising directions for future research.

## Methods

### Problem Statement

In this paper, we mainly focused on the traditional CIL scenario and the emerging streaming learning scenario. In CIL, we denote sequentially arriving tasks as $D = \{D_1, \ldots, D_T\}$ where each task $D_t = \{(\mathbf{x}_i^t, y_i^t)\}_{i=1}^{N_t}$ contains $N_t$ samples. Each sample $\mathbf{x}_i^t$ within a task is drawn from the input space $X_t$, and its corresponding label $y_i^t$ from the label space $Y_t$. The training process involves sequentially learning from $D_1$ to $D_T$, followed by class prediction on an unseen test set within the full label space. The neural network learns mutually exclusive classes within each task where the intersection of $Y_i \cap Y_j = \emptyset$. For streaming learning, the notation is similar, but the task sequence is much longer, and samples within each task can only be trained once. In this work, we primarily focused on the case where $Y_i = Y_j$ to study plasticity changes throughout the process.

### Dataset Configuration

In Fig. 2, we generated simulated data to mimic odors using a configuration similar to previous work [49]. We first generated $C = 10$ class prototypes, $\mathbf{c_i} \in \mathbf{R}^n$, from a uniform distribution $U[0,1]^n$. Each prototype is a $n = 50$ dimensional vector. Subsequently, we applied Gaussian noise to generate samples $\mathbf{x_{ij}} = \mathbf{c_i} + N(\mathbf{0}, \mathbf{\Sigma})$, where each sample is labeled with $y_i$ corresponding to the category of $\mathbf{c_i}$. For simplicity, we assume different dimensions of the Gaussian noise are independent and identically distributed, which means the covariance matrix $\mathbf{\Sigma} = \text{diag}(\sigma^2, \ldots, \sigma^2)$. For CL, we adopted the CIL setting with task sequence $T = 5$, and each task has 2 classes. We set the standard deviation of the Gaussian noise to $\sigma = 0.5$. There are a total of 50,000 training samples and 10,000 testing samples, with the number of samples balanced across classes.

In Figs. 4-5 and Supplementary Fig. 2, we employed the same experimental setting. For CIFAR-100, the task sequence $T = 10$, with each task comprising 10 randomly selected classes. For CUB-200-2011, the task sequence $T = 10$, and each task has 20 classes. For VTAB, the task sequence $T = 5$, and each task has 10 classes.

In Fig. 6, we used Input Permuted MNIST and Label Permuted MNIST. Similar to [29], we randomly sampled 10,000 samples from the raw MNIST dataset and divided them into different batches. For different tasks, we generated different permutation orders and shuffled the data (images for Input Permuted MNIST and labels for Label Permuted MNIST) before sending them to the neural network. We conducted the experiment with a task sequence $T = 100$.

In Supplementary Fig. 3, we used the same configuration as in Fig. 4, except that the samples across different classes are imbalanced. The number of samples in class $k$ is denoted as $n_k$, and $n_i \neq n_j$ when $i \neq j$. We assume $\{n_k\}$ is sorted by cardinality in descending order, which means $n_1 > n_2 > \cdots > n_C$. In class-imbalanced setting, we define the imbalance ratio as $\gamma = \frac{\max\{n_k\}}{\min\{n_k\}}$ to measure the degree of class imbalance. In the *Normal* setup, $n_k$ follows an exponential distribution given by $n_k = n_1 \gamma^{-\frac{k-1}{C-1}}$. This is reversed in *Reverse* setup and randomly shuffled in *Random* setup.

**Mathematical Formula**

We traced the process of forward propagation and backward propagation to demonstrate how the Fly Model operates from a mathematical perspective. Assume we have a feature $\mathbf{x} \in \mathbf{R}^n$ input to the Fly Model, it is then transformed to a higher dimension feature $\mathbf{h} \in \mathbf{R}^m$ where $m \gg n$ using a binary, sparse random matrix $W_{P \to K} \in R^{m \times n}$ through $\mathbf{h} = W_{P \to K}\mathbf{x}$. In each row of $W_{P \to K}$, $k \ll n$ random columns are set to 1, while the remaining columns are set to 0. During the learning process, $W_{P \to K}$ is not updated. The high-dimensional feature $\mathbf{h}$ is then inhibited by a top-$k$ operation $top\text{-}k(\mathbf{x}) : \mathbf{R}^m \to \mathbf{R}^m$, with only $m \times k \ll m$ dimensions with the strongest activations are retained, and the remaining dimensions are reset to 0. This can be formulated as $\mathbf{z} \in R^m$ satisfying $\mathbf{z} = top\text{-}k(\mathbf{h})$. After the top-$k$ operation, $\mathbf{z}$ is projected to output neurons for classification $\mathbf{o} = W_{K \to M}\mathbf{z}$, where $\mathbf{o} \in \mathbf{R}^c$ and $W_{K \to M} \in \mathbf{R}^{c \times m}$. Unlike $W_{P \to K}$, $W_{K \to M}$ is updated through backpropagation. Assume the loss function is $L(\mathbf{o}, y)$, $y$ is the ground truth label, the synaptic weights in $W_{K \to M}$ are updated by:

$$W_{ij} \leftarrow W_{ij} - learning\_rate \times \frac{\partial L(o,y)}{\partial W_{ij}},$$

where

$$\frac{\partial L(\mathbf{o}, y)}{\partial W_{ij}} = \begin{cases} \frac{\partial L(\mathbf{o}, y)}{\partial W_{ij}}, & \text{if the j-th dimension is activated in z,} \\ 0, & \text{if the j-th dimension is not activated in z.} \end{cases}$$

This means that most parameters in $W_{K \to M}$ are not updated.

**Angle Between Two Random Vectors**

In an *n*-dimensional space, the probability density function of the angle between two random vectors is:

$$p_n(\theta) = \frac{\Gamma(\frac{n}{2})}{\Gamma(\frac{n-1}{2})\sqrt{\pi}} \sin^{n-1}\theta.$$

The variance of the angle distributions is:

$$Var_n(\theta) = \frac{\Gamma(\frac{n}{2})}{\Gamma(\frac{n-1}{2})\sqrt{\pi}} \int_0^\pi (\theta - \frac{\pi}{2})^2 \sin^{n-2}\theta d\theta,$$

which decreases as *n* increases. Thus, higher dimensions increase the probability that two vectors become orthogonal.

**Probability of Each KC Connecting to Different Set of PNs**

This is a variant of the well-known "birthday problem" [46], which can be calculated using combinatorial numbers. We assume that the PN layer has *n* neurons, the KC layer has *m* neurons and the synaptic degree is *r*. We can denote $R = C_n^r$. Therefore, the probability is given by:

$$p = \frac{R!}{(R-m)! R^m}.$$

**More Orthogonal Gradients Have Less Forgetting**

We denote $\nabla L_t(w)$ as gradient at task *t*. Then we can derive the following theorem:

**Theorem 1** [52]: *Let $w' = w - \eta \nabla L_t(w)$, Then there exists $\xi \in [0,1]$ such that $\nabla L_1(w) - \nabla L_t(w') = -\eta < \nabla L_1(w - \xi \eta \nabla L_t(w)), \nabla L_t(w) >$.*

Obviously, the left side of the equation increases, leading to forgetting. The right side of the equation represents the inner product of the two gradients. Therefore, if the two gradients are more orthogonal to each other, the inner product decreases, and the neural network forgets less about the previous tasks.

**Network Architecture**

For the pre-trained model Vision Transformer, we used the Vit B/16-IN21K version across all experiments. We integrated the Fly Model with several multilayer perceptrons for testing.
- In Figs. 2, 4-5 and Supplementary Fig. 3, we used the simplest Fly Model consisting of three layers.
- In Fig. 6a, we prepended two linear layers to the Fly Model, with a hidden dimension of 784 (matching the dimension of MNIST images) and ReLU as the activation function between these layers.
- In Fig. 6b, we added one linear layer compared to the network in Fig. 4, with the hidden dimension set to 100.
- In Supplementary Fig. 2, we used the same architecture as Fig. 6a, but the hidden dimension is 768 (the same as the output dimension of the Vision Transformer).

**Compared Baseline**

Apart from the original SGD, we choose two forgetting-targeted methods EWC, SI and three plasticity-targeted methods Shrink and Perturb (S&P), L2 Init, and Continual Backpropagation (CBP) for comparison.

Elastic Weight Consolidation (EWC) [6] is one of the most classical CL methods. It uses the Fisher information matrix to record the weights important for old tasks and protects them through a regularization term during parameter updates. Let $L_{train}(\theta)$ be the loss function for new tasks, $F$ be the Fisher information matrix, $\lambda$ be the regularization coefficient, $\theta_i$ and $\theta^*_{1:t-1,i}$ be the current and previous parameters, respectively. The EWC loss is defined as:

$$L_{EWC}(\theta) = L_{train}(\theta) + \frac{\lambda}{2} \sum_i F_{1:t-1,i}(\theta_i - \theta^*_{1:t-1,i})^2.$$

Synaptic Intelligence (SI) [9] is similar to EWC but has lower complexity, reducing the task of maintaining an $O(n^2)$ Fisher information matrix to $O(n)$ importance weights for each parameter. Additionally, it has better biological interpretability. The SI loss is defined as:

$$L_{SI}(\theta) = L_{train}(\theta) + c \sum_k \Omega_k^\mu (\theta_k - \theta_k^*)^2,$$

where $\Omega_k^\mu = \sum_{v<\mu} \frac{\omega_k^v}{(\Delta_k^v)^2 + \xi}$ denotes the importance of the parameter. $c$ is the regularization coefficient.

Shrink and Perturb (S&P) [28] imposes perturbations by first slightly shrinking the network's weights and then adding a small-scale perturbation from randomly initialized parameter at the end of each task. Here, *shrink* and *perturb* are small numeric numbers. The modified weight is:

$$w_i' = (1 - shrink) \times w_i + perturb \times w_i^0.$$

L2 Init [29] imposes a regularization term on the network weights during the training process, constraining their distance from the randomly initialized weights. Let $L_{train}(\theta)$ be the loss function of the task, $\theta_0$ be the initial parameters, and $\alpha$ be the regularization coefficient. The regularized loss is:

$$L_{reg}(\theta) = L_{train}(\theta) + \alpha ||\theta - \theta_0||_2^2.$$

Continual Backpropagation (CBP) [5] tracks the utility measure for each neuron and periodically reinitializes the subset of weights that have the smallest utility values. The utility is calculated as:

$$u_l[i] = \eta \times u_l[i] + (1 - \eta) \times |h_{l,i,t}| \times \sum_{k=1}^{n_{l+1}} |w_{l,i,k,t}|,$$

where $h_{l,i,t}$ is the output of the $i$-th neuron in in layer $l$ at time $t$, and $w_{l,i,k,t}$ is the weight connecting the $i$-th neuron of layer $l$ and the $k$-th neuron of layer $l+1$ at time $t+1$.

**Evaluation Metric**

To evaluate performance under CIL setting, we mainly used the following metrics: average accuracy ($A_t$), accumulated accuracy ($\bar{A}_t$), backward transfer ($BWT_t$) and forward transfer ($FWT_t$)

$$A_t = \frac{1}{t} \sum_{i=1}^{t} a_{t,i},$$

$$\bar{A}_t = \frac{1}{t} \sum_{i=1}^{t} A_t,$$

$$BWT_t = \frac{1}{t-1} \sum_{i=1}^{t} (a_{t,i} - a_{i,i}),$$

$$FWT_t = \frac{1}{t-1} \sum_{i=2}^{t} (a_{i,i} - \tilde{a}_i),$$

where $a_{t,i}$ is the test accuracy of task $i$ after learning task $t$, and $\tilde{a}_i$ is the test accuracy of task $i$ when learned directly from a randomly initialized network.

Throughout the process, $A_t$ can be viewed as the performance of CL at each stage. $\bar{A}_t$ can be regarded as the average accumulated performance before task $t$. $BWT_t$ measures how much of the previously learned tasks is forgotten after learning task t, representing the degree of catastrophic forgetting. $FWT_t$ measures the performance decrement for each newly learned class relative to learning from scratch, which indicates plasticity loss.

For the streaming learning setting, we mainly focused on plasticity loss. This can be measured by average online accuracy, dormant units, stable rank, and average weight magnitude. Their specific forms are listed below.

$$average\ online\ accuracy = \frac{1}{N_i} \sum_{j=1}^{N_i} a_{i,j},$$

where $N_i$ is the number of data batches in task $i$, and $a_{i,j}$ is the average accuracy of the $j$-th batch.

$$dormant\ units = \frac{1}{M} \sum_{i=1}^{M} I(a_i < \delta),$$

where $a_i$ is the activation value of each neuron in the network (with $M$ neurons), and $\delta = 0.01$ is the threshold. Smaller activation values can lead to gradients tending toward zero during backpropagation, making it difficult for the weights to be updated and results in plasticity loss.

$$stable\ rank = \min_{k} \frac{\sum_{i=1}^{k} \sigma_i}{\sum_{i=1}^{n} \sigma_n} \geq 1 - \delta,$$

where $\sigma_i$ are singular values of the preceding linear layer of the Fly Model, sorted as $\sigma_1 > \sigma_2 > \cdots > \sigma_n$. Following [29, 56], we use $\delta = 0.01$ as threshold. A low stable rank indicates that only a small number of dimensions are needed to represent most of the information in the matrix, suggesting that many units in the hidden layer do not provide effective output, which is a sign of plasticity loss.

$$average\ weight\ magnitude = \frac{1}{M} \sum_{i=1}^{M} w_i,$$

where $w_i$ is the weight value of each neuron (with $M$ neurons). The magnitude of the network's weights is related to the condition number of the Hessian matrix in the second-order Taylor expansion of the loss function. An increase in network weights may lead to a pathological Hessian matrix. According to convex optimization theory, this can slow down the convergence speed of the SGD algorithm, requiring a longer time for the neural network to converge [53].

**Implementation Detail**

We used grid search to find the optimal hyperparameters for each experiment. The best parameters are summarized in Supplementary Table 1. All results were averaged over 5 runs using different random seeds. Additionally, we used gradient clipping to avoid gradient explosion if necessary.

**Data availability**

All datasets analyzed in this study are freely available online resources.

**Code availability**

Code that can be used to reproduce the main results of this paper will be available on github.

**References**

[1] Van de Ven, G. M., Tuytelaars, T. & Tolias, A. S. Three types of incremental learning. *Nat. Mach. Intell.* **4**, 1185–1197 (2022).

[2] Van de Ven, G. M., & Tolias, A. S. Three scenarios for continual learning. Preprint at https://arxiv.org/abs/1904.07734 (2019).

[3] Wang, L., Zhang, X., Su, H. & Zhu, J. A comprehensive survey of continual learning: theory, method and application. *IEEE Trans. Pattern Anal. Mach. Intell.* **46**, 5362–5383 (2024).

[4] Lyle, C., Zheng, Z., Nikishin, E., Pires, B. A., Pascanu, R., & Dabney, W. Understanding plasticity in neural networks. In *Proc. International Conference on Machine Learning* 23190-23211 (PMLR, 2023).

[5] Dohare, S., Hernandez-Garcia, J. F., Lan, Q., Rahman, P., Mahmood, A. R., & Sutton, R. S. Loss of plasticity in deep continual learning. *Nature* **632**, 768-774 (2024).

[6] Kirkpatrick, J., Pascanu, R., Rabinowitz, N., Veness, J., Desjardins, G. et al. Overcoming catastrophic forgetting in neural networks. *Proc. Natl Acad. Sci. USA* **114**, 3521–3526 (2017).

[7] Li, Z., & Hoiem, D. Learning without forgetting. *IEEE Trans. Pattern Anal. Mach. Intell.* **40**, 2935-2947 (2017).

[8] Aljundi, R., Babiloni, F., Elhoseiny, M., Rohrbach, M., & Tuytelaars, T. Memory aware synapses: learning what (not) to forget. In *Proc. European Conference on Computer Vision* 139–154 (Springer, 2018).

[9] Zenke, F., Poole, B., & Ganguli, S. Continual learning through synaptic intelligence. In *Proc. International Conference on Machine Learning* 3987–3995 (PMLR, 2017).

[10] Schwarz, J., Czarnecki, W., Luketina, J., Grabska-Barwinska, A., Teh, Y. W., Pascanu, R., & Hadsell, R. Progress & compress: A scalable framework for continual learning. In *Proc. International Conference on Machine Learning* 4528-4537 (PMLR, 2018).

[11] Ahn, H., Cha, S., Lee, D., & Moon, T. Uncertainty-based continual learning with adaptive regularization. *Adv. Neural Inf. Process. Syst.* **32**, 4392-4402 (2019).

[12] Serra, J., Suris, D., Miron, M., & Karatzoglou, A. Overcoming catastrophic forgetting with hard attention to the task. In *Proc. International Conference on Machine Learning* 4548-4557 (PMLR, 2018).

[13] Zeng, G., Chen, Y., Cui, B., & Yu, S. Continual learning of context-dependent processing in neural networks. *Nat. Mach. Intell.* **1**, 364-372 (2019).

[14] Mallya, A., Davis, D., & Lazebnik, S. Piggyback: Adapting a single network to multiple tasks by learning to mask weights. In *Proc. European Conference on Computer Vision* 67-82 (Springer, 2018).

[15] Mallya, A., & Lazebnik, S. Packnet: Adding multiple tasks to a single network by iterative pruning. In *Proc. IEEE Conference on Computer Vision and Pattern Recognition* 7765-7773 (IEEE, 2018).

[16] Rusu, A. A., Rabinowitz, N. C., Desjardins, G., Soyer, H., Kirkpatrick, J. et al. Progressive neural networks. Preprint at https://arxiv.org/abs/1606.04671 (2016).

[17] Fernando, C., Banarse, D., Blundell, C., Zwols, Y., Ha, D. et al. Pathnet: Evolution channels gradient descent in super neural networks. Preprint at https://arxiv.org/abs/1701.08734 (2017).

[18] Lopez-Paz, D., & Ranzato, M. A. Gradient episodic memory for continual learning. *Adv. Neural Inf. Process. Syst.* **30**, 6467-6476 (2017).

[19] Rebuffi, S. A., Kolesnikov, A., Sperl, G., & Lampert, C. H. icarl: Incremental classifier and representation learning. In *Proc. IEEE Conference on Computer Vision and Pattern Recognition* 2001-2010 (IEEE, 2017).

[20] Aljundi, R., Belilovsky, E., Tuytelaars, T., Charlin, L., Caccia, M. et al. Online continual learning with maximal interfered retrieval. *Adv. Neural Inf. Process. Syst.* **32**, 11849-11860 (2019).

[21] Wang, F. Y., Zhou, D. W., Ye, H. J., & Zhan, D. C. Foster: Feature boosting and compression for class-incremental learning. In *Proc. European Conference on Computer Vision* 398-414 (Springer, 2022).

[22] van de Ven, G.M., Siegelmann, H.T. & Tolias, A.S. Brain-inspired replay for continual learning with artificial neural networks. *Nat. Commun.* **11**, 4069 (2020).

[23] Liu, X., Wu, C., Menta, M., Herranz, L., Raducanu, B. et al. Generative feature replay for class-incremental learning. In *Proc. IEEE Conference on Computer Vision and Pattern Recognition Workshops* 226-227 (IEEE, 2020).

[24] Nikishin, E., Schwarzer, M., D'Oro, P., Bacon, P. L., & Courville, A. The primacy bias in deep reinforcement learning. In *Proc. International Conference on Machine Learning* 16828-16847 (PMLR, 2022).

[25] Nikishin, E., Oh, J., Ostrovski, G., Lyle, C., Pascanu, R. et al. Deep reinforcement learning with plasticity injection. *Adv. Neural Inf. Process. Syst.* **36**, 37142-37159 (2024).

[26] Sokar, G., Agarwal, R., Castro, P. S., & Evci, U. The dormant neuron phenomenon in deep reinforcement learning. In *Proc. International Conference on Machine Learning* 32145-32168 (PMLR, 2023).

[27] Igl, M., Farquhar, G., Luketina, J., Boehmer, W., & Whiteson, S. Transient non-stationarity and generalisation in deep reinforcement learning. In *International Conference on Learning Representations* (2021).

[28] Ash, J., & Adams, R. P. On warm-starting neural network training. *Adv. Neural Inf. Process. Syst.* **33**, 3884-3894 (2020).

[29] Kumar, S., Marklund, H., & Van Roy, B. Maintaining Plasticity in Continual Learning via Regenerative Regularization. In *Proc. 3rd Conference on Lifelong Learning Agents* (PMLR, 2024).

[30] Zilly, J., Achille, A., Censi, A., & Frazzoli, E. On plasticity, invariance, and mutually frozen weights in sequential task learning. *Adv. Neural Inf. Process. Syst.* **34**, 12386-12399 (2021).

[31] Lyle, C., Rowland, M., & Dabney, W. Understanding and preventing capacity loss in reinforcement learning. In *International Conference on Learning Representations* (2022).

[32] Abbas, Z., Zhao, R., Modayil, J., White, A., & Machado, M. C. Loss of plasticity in continual deep reinforcement learning. In *Conference on Lifelong Learning Agents* 620-636 (PMLR, 2023).

[33] Shang, W., Sohn, K., Almeida, D., & Lee, H. Understanding and improving convolutional neural networks via concatenated rectified linear units. In *Proc. International Conference on Machine Learning* 2217-2225 (PMLR, 2016).

[34] Chaudhry, A., Dokania, P. K., Ajanthan, T., & Torr, P. H. Riemannian walk for incremental learning: Understanding forgetting and intransigence. In *Proc. European Conference on Computer Vision* 532-547 (Springer, 2018).

[35] Kim, S., Noci, L., Orvieto, A., & Hofmann, T. Achieving a better stability-plasticity trade-off via auxiliary networks in continual learning. In *Proc. IEEE Conference on Computer Vision and Pattern Recognition* 11930-11939 (IEEE, 2023).

[36] Elsayed, M., & Mahmood, A. R. Addressing loss of plasticity and catastrophic forgetting in continual learning. In *International Conference on Learning Representations* (2024).

[37] Wang, L., Zhang, X., Li, Q., Zhang, M., Su, H. et al. Incorporating neuro-inspired adaptability for continual learning in artificial intelligence. *Nat. Mach. Intell.* **5**, 1356-1368 (2023).

[38] Luo, L. Architectures of neuronal circuits. *Science* **373**, eabg7285 (2021).

[39] Lin, A. C., Bygrave, A. M., De Calignon, A., Lee, T., & Miesenböck, G. Sparse, decorrelated odor coding in the mushroom body enhances learned odor discrimination. *Nat. Neurosci.* **17**, 559-568 (2014).

[40] Stevens, C. F. What the fly's nose tells the fly's brain. *Proc. Natl Acad. Sci. USA* **112**, 9460-9465 (2015).

[41] Caron, S. J., Ruta, V., Abbott, L. F., & Axel, R. Random convergence of olfactory inputs in the Drosophila mushroom body. *Nature* **497**, 113-117 (2013).

[42] Dasgupta, S., Stevens, C. F., & Navlakha, S. A neural algorithm for a fundamental computing problem. *Science* **358**, 793-796 (2017).

[43] Krizhevsky, A., & Hinton, G. Learning multiple layers of features from tiny images. *Technical Report* (2009).

[44] Wah, C., Branson, S., Welinder, P., Perona, P., & Belongie, S. The caltech-ucsd birds-200-2011 dataset. *Technical Report* (2011).

[45] Zhai, X., Puigcerver, J., Kolesnikov, A., Ruyssen, P., Riquelme, C. et al. A large-scale study of representation learning with the visual task adaptation benchmark. Preprint at https://arxiv.org/abs/1910.04867 (2019).

[46] Litwin-Kumar, A., Harris, K. D., Axel, R., Sompolinsky, H., & Abbott, L. F. Optimal degrees of synaptic connectivity. *Neuron* **93**, 1153-1164 (2017).

[47] Babadi, B., & Sompolinsky, H. Sparseness and expansion in sensory representations. *Neuron* **83**, 1213-1226 (2014).

[48] Brunel, N., Hakim, V., Isope, P., Nadal, J. P., & Barbour, B. Optimal information storage and the distribution of synaptic weights: perceptron versus Purkinje cell. *Neuron* **43**, 745-757 (2004).

[49] Wang, P. Y., Sun, Y., Axel, R., Abbott, L. F., & Yang, G. R. Evolving the olfactory system with machine learning. *Neuron* **109**, 3879-3892 (2021).

[50] Turner, G. C., Bazhenov, M., & Laurent, G. Olfactory representations by Drosophila mushroom body neurons. *J. Neurophysiol.* **99**, 734-746 (2008).

[51] Campbell, R. A., Honegger, K. S., Qin, H., Li, W., Demir, E., & Turner, G. C. Imaging a population code for odor identity in the Drosophila mushroom body. *J Neurosci.* **33**, 10568-10581 (2013).

[52] Mirzadeh, S. I., Chaudhry, A., Yin, D., Hu, H., Pascanu, R. et al. Wide neural networks forget less catastrophically. In *Proc. International Conference on Machine Learning* 15699-15717 (PMLR, 2022).

[53] Boyd, S., & Vandenberghe, L. *Convex optimization* (Cambridge Univ. Press, 2004).

[54] Dosovitskiy, A, Beyer, L., Kolesnikov, A., Weissenborn, D., Zhai, X. et al. An image is worth 16x16 words: Transformers for image recognition at scale. In *International Conference on Learning Representations* (2021).

[55] Goodfellow, I. J., Mirza, M., Xiao, D., Courville, A., & Bengio, Y. An empirical investigation of catastrophic forgetting in gradient-based neural networks. Preprint at https://arxiv.org/abs/1312.6211 (2013).

[56] Kumar, A., Agarwal, R., Ghosh, D., & Levine, S. Implicit under-parameterization inhibits data-efficient deep reinforcement learning. In *International Conference on Learning Representations* (2021).

[57] Imam, N., & Cleland, T. A. Rapid online learning and robust recall in a neuromorphic olfactory circuit. *Nat. Mach. Intell.* **2**, 181-191 (2020).

[58] Papadopoulou, M., Cassenaer, S., Nowotny, T., & Laurent, G. Normalization for sparse encoding of odors by a wide-field interneuron. *Science* **332**, 721-725 (2011).

[59] Zang, Y., & De Schutter, E. Recent data on the cerebellum require new models and theories. *Curr. Opin. Neurobiol.* **82**, 102765 (2023).

[60] Westeinde, E. A., Kellogg, E., Dawson, P. M., Lu, J., Hamburg, L. et al. Transforming a head direction signal into a goal-oriented steering command. *Nature* **626**, 819-826 (2024).

[61] Zang, Y., & De Schutter, E. The cellular electrophysiological properties underlying multiplexed coding in Purkinje cells. *J Neurosci.* **41**, 1850-1863 (2021).

[62] Zang, Y., Dieudonné, S., & De Schutter, E. Voltage-and branch-specific climbing fiber responses in Purkinje cells. *Cell reports* **24**, 1536-1549 (2018).

[63] Zang, Y., & De Schutter, E. Climbing fibers provide graded error signals in cerebellar learning. *Front. Syst. Neurosci.* **13**, 1662-5137 (2019).

[64] Zang, Y., Marder, E., & Marom, S. Sodium channel slow inactivation normalizes firing in axons with uneven conductance distributions. *Current biology* **33**, 1818-1824 (2023).

[65] Wu, Y., Huang, L. K., Wang, R., Meng, D., & Wei, Y. Meta Continual Learning Revisited: Implicitly Enhancing Online Hessian Approximation via Variance Reduction. In *International Conference on Learning Representations* (2024).

# Structural features of the fly olfactory circuit mitigate the stability-plasticity dilemma in continual learning

## Additional Discussion

There are approaches that aim to simultaneously improve stability and plasticity compared to traditionally used methods, such as RWalk [34], ANCL [35], CAF [37], and UPGD [36]. For RWalk, ANCL, and CAF, these methods introduce an additional regularization term specifically designed to mitigate plasticity loss within stability-targeted frameworks. By tuning two penalty coefficients (one for stability and one for plasticity), they achieve a better balance. UPGD approaches the problem from an optimization perspective, adding constraints for old tasks to maintain stability while introducing perturbations to preserve plasticity.

Their strategies are similar to finding the optimal coding level in the KC layer to balance stability and plasticity (see our Fig. 3f). Nonetheless, the representational capacity remains fixed. The only way to push the boundary further is by introducing more parameters, which will inevitably result in significantly higher computational costs. The value of the Fly Model lies in its biological design, which introduces many more parameters while sparse connections and sparse coding minimize additional computational costs (Fig. 3g). This provides an elegant and efficient way to truly resolve the stability-plasticity dilemma. To the best of our knowledge, we are the first to achieve this in a biologically plausible manner.

## Additional Results

**The Fly Model facilitates continual learning in multilayer networks.** We have tested the Fly Model's effects by directly combining it with a pre-trained model. To assess the Fly Model's generality, we evaluated its impact in a more complex configuration by replacing the classification head of a multilayer network with the Fly Model (Supplementary Fig. 2a). All other configurations match those in Figs. 4-5. All synaptic weights are plastic except in the pre-trained model and the Fly Model's expansion layer. Additionally, a synaptic update strategy aimed at mitigating plasticity loss in multilayer networks was tested [5].

Consistent with results in Figs. 4-5, introducing the Fly Model in a multilayer network reliably improves accuracy across all datasets and synaptic update methods (Supplementary Fig. 2b-c). In Fig. 4, EWC and SI show significant improvement over SGD with a two-layer perceptron after the pre-trained model. This improvement diminishes when multilayer networks follow the pre-trained model (Supplementary Fig. 2b, SGD, EWC, and SI). Nonetheless, using the Fly Model significantly enhances test accuracy, with SI showing the largest increase. For CIFAR-100, the average accuracy increases by 9% (SGD), 11% (EWC), and 15% (SI); for CUB-200-2011, it increases by 32% (SGD), 34% (EWC), and 44% (SI). The performance improvement for VTAB is comparable to that for CIFAR-100. Plasticity-targeted methods exhibit performance similar to SGD, but combining them with the Fly Model significantly improves performance (Supplementary Fig. 2c).

**The reliability of the Fly Model in class-imbalanced scenarios.** In real-world applications, data often come from imbalanced sampling, leading to significantly varied numbers of samples across different tasks in continual learning. This imbalance can cause performance drops in ML algorithms. To test the reliability of the Fly Model against imbalanced data, we simulated three scenarios in Supplementary Fig. 3: Normal, Reverse, and Random [65]. In the Normal scenario, the number of samples in each class decreases throughout the learning process. The Reverse scenario shows the opposite order. In the Random scenario, the sample numbers of different classes are randomly distributed. We tested three values of $\gamma$, which determine the imbalance ratio (see Methods). A larger $\gamma$ indicates more imbalance in the data samples for each task. The simulations were performed on CIFAR-100 under the same CIL setting as in Fig. 4. Across different class-imbalanced scenarios, the Fly Model consistently improves test accuracy over the baseline model for all tested synaptic update strategies and imbalance ratios. These results indicate that the Fly Model can serve as a reliable module for continual learning with imbalanced data in future real-world applications.

**Supplementary Table 1: Detailed hyperparameter configuration for each experiment.** lr denotes the learning rate. PN, KC, $r$, and $k$ define the structure of the Fly Model. $\lambda$, c, and $\alpha$ are the regularization coefficients corresponding to EWC, SI and L2_Init, respectively. *shrink* and *perturb* correspond to the Shrink & Perturb methods, respectively. For datasets, "Input" and "Label" are abbreviated for "Input Permuted MNIST" and "Label Permuted MNIST," respectively. Abbreviation like "Normal-2" refers to datasets using CIFAR-100 in "Normal" imbalanced setup with an imbalanced ratio $\gamma = 2$. A blank block indicates that the parameter is not involved.

| Dataset | lr | PN | KC | $r$ | $k$ | $\lambda$ | c | epoch | bs | $\alpha$ | *shrink* | *perturb* |
|---|---|---|---|---|---|---|---|---|---|---|---|---|
| Fig. 2 | | | | | | | | | | | | |
| Odor | 0.005 | 50 | 2000 | 6 | 0.01 | | 5 | 1 | 64 | | | |
| Fig. 4 | | | | | | | | | | | | |
| CIFAR-100 | 0.001 | 768 | 30000 | 40 | 0.001 | 5000 | 1000 | 5 | 128 | | | |
| CUB-200 | 0.004 | 768 | 30000 | 40 | 0.001 | 500 | 1000 | 5 | 128 | | | |
| VTAB | 0.01 | 768 | 30000 | 40 | 0.005 | 50 | 50 | 5 | 128 | | | |
| Fig. 5 | | | | | | | | | | | | |
| CIFAR-100 | 0.001 | 768 | 30000 | 40 | 0.001 | | | 5 | 128 | 1e-5 | 1e-4 | 1e-3 |
| CUB-200 | 0.004 | 768 | 30000 | 40 | 0.001 | | | 5 | 128 | 1e-5 | 1e-4 | 1e-3 |
| VTAB | 0.01 | 768 | 30000 | 40 | 0.005 | | | 5 | 128 | 1e-5 | 1e-4 | 1e-3 |
| Fig. 6 | | | | | | | | | | | | |
| Input | 0.05 | 784 | 30000 | 40 | 0.001 | 10 | 0.01 | 1 | 100 | | | |
| Label | 0.01 | 784 | 30000 | 40 | 0.001 | 1 | 0.1 | 1 | 100 | | | |
| Supplementary Fig. 2 | | | | | | | | | | | | |
| CIFAR-100 | 0.005 | 768 | 30000 | 80 | 0.01 | 50 | 100 | 5 | 128 | 1e-5 | 1e-4 | 1e-3 |
| CUB-200 | 0.02 | 768 | 30000 | 80 | 0.01 | 10 | 20 | 5 | 128 | 1e-5 | 1e-4 | 1e-3 |
| VTAB | 0.05 | 768 | 30000 | 40 | 0.005 | 10 | 20 | 5 | 128 | 1e-5 | 1e-4 | 1e-3 |
| Supplementary Fig. 3 | | | | | | | | | | | | |
| Normal-2 | 0.002 | 768 | 30000 | 40 | 0.001 | 2000 | 1000 | 5 | 128 | | | |
| Normal-5 | 0.003 | 768 | 30000 | 40 | 0.001 | 500 | 1000 | 5 | 128 | | | |
| Normal-10 | 0.005 | 768 | 30000 | 40 | 0.001 | 50 | 10 | 5 | 128 | | | |
| Reverse-2 | 0.002 | 768 | 30000 | 40 | 0.001 | 2000 | 1000 | 5 | 128 | | | |
| Reverse-5 | 0.003 | 768 | 30000 | 40 | 0.001 | 500 | 1000 | 5 | 128 | | | |
| Reverse-10 | 0.005 | 768 | 30000 | 40 | 0.001 | 50 | 10 | 5 | 128 | | | |
| Random-2 | 0.002 | 768 | 30000 | 40 | 0.001 | 5000 | 1000 | 5 | 128 | | | |
| Random-5 | 0.003 | 768 | 30000 | 40 | 0.001 | 500 | 1000 | 5 | 128 | | | |
| Random-10 | 0.005 | 768 | 30000 | 40 | 0.001 | 500 | 100 | 5 | 128 | | | |

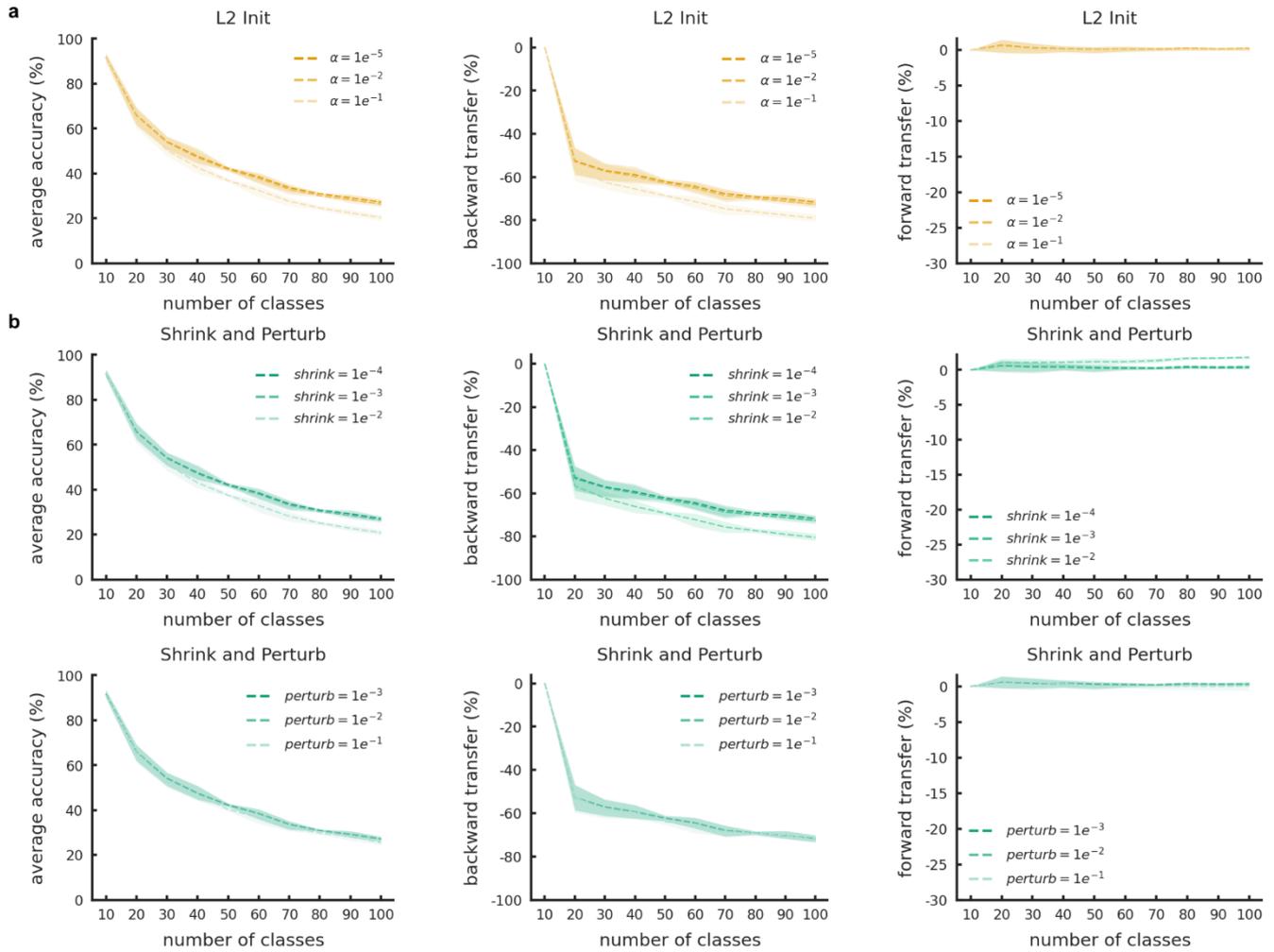

**Supplementary Fig.1 Large plasticity-targeted hyperparameter will deteriorate accuracy. a** Influence of different $\alpha$ in L2 Init. **b** Influence of different *shrink* and *perturb* in Shrink and Perturb. Both **a** and **b** were evaluated on CIFAR-100 using the same configuration as in Fig. 5. $\alpha$ in L2 Init and *shrink* in Shrink and Perturb have a significant impact on accuracy, while *perturb* in Shrink and Perturb shows no noticeable effect。

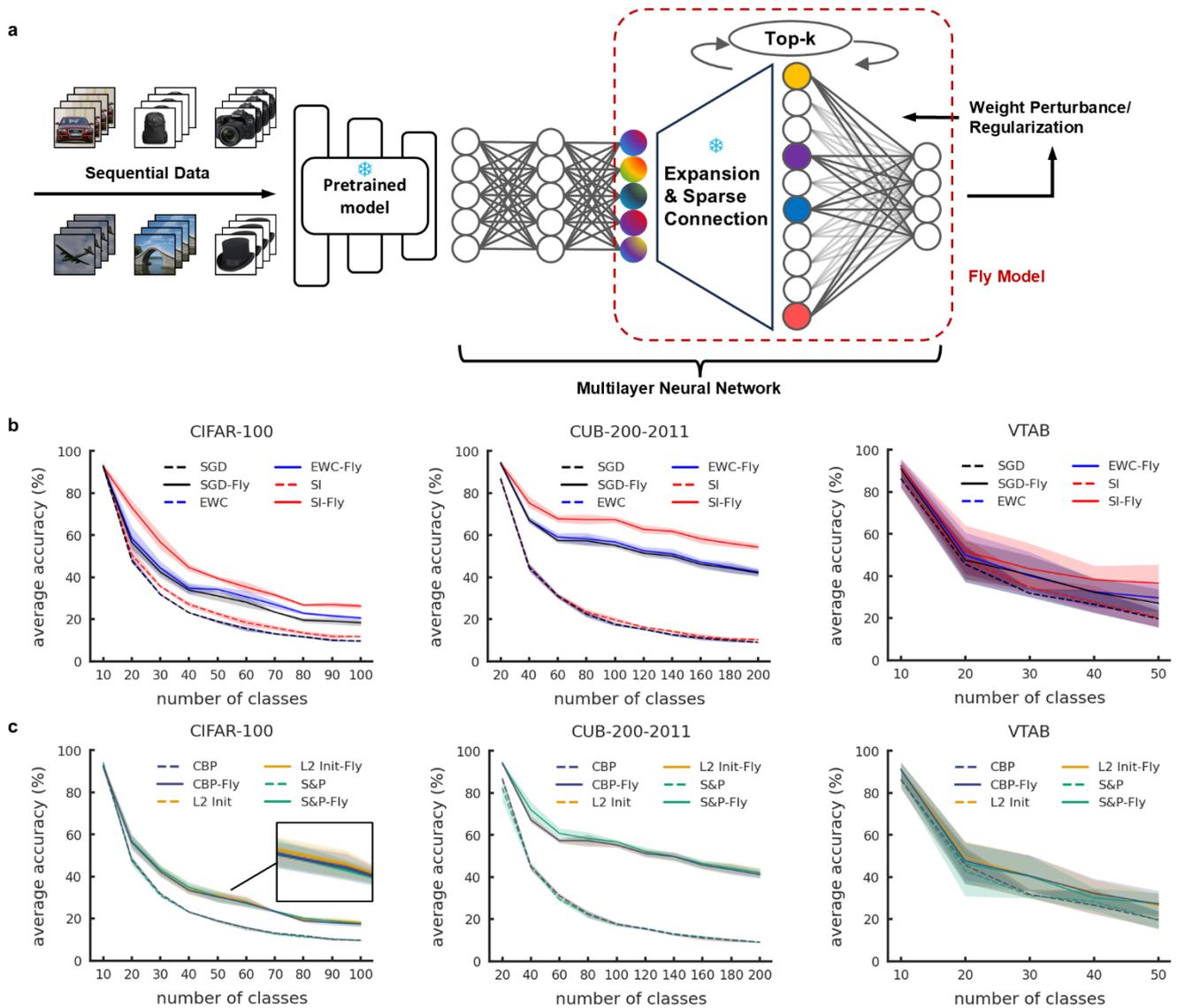

**Supplementary Fig.2 The Fly Model promotes continual learning in multilayer networks. a** Model pipeline. A multilayer neural network is placed between the Fly Model and the pre-trained layer. **b** Average accuracy, backward transfer, and forward transfer against the number of classes learned using CIFAR-100, CUB-200-2011, and VTAB. SGD (-Fly), EWC (-Fly), and SI (-Fly) refer to the meanings in Fig. 4. **c** Same as **b**, but with plasticity-targeted methods CBP, L2 Init, and S&P included.

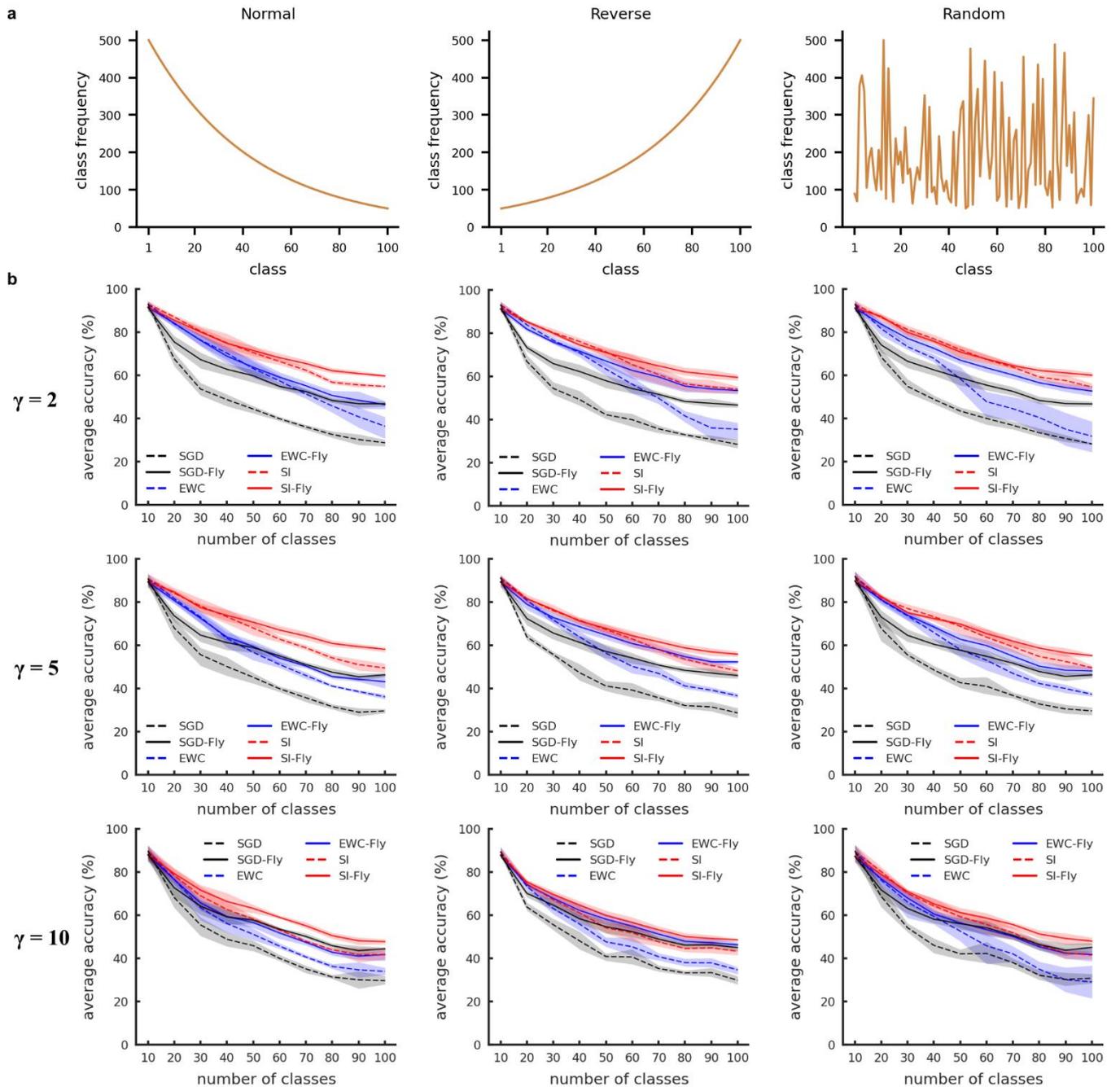

**Supplementary Fig. 3 The Fly Model reliably enhances continual learning under class-imbalanced scenarios. a** Sample numbers per category across three types of imbalance paradigms. **b** Test accuracy of the Fly Model across three class-imbalanced scenarios (Normal, Reverse, Random, from left to right). Three imbalance ratios are used ($\gamma$ =2, 5, 10 from top to bottom). Synaptic update strategies include SGD, EWC, and SI.